\documentclass[runningheads]{llncs}
\usepackage[T1]{fontenc}
\usepackage{multirow}
\usepackage[normalem]{ulem}
\useunder{\uline}{\ul}{}
\usepackage{amsmath}
\usepackage{amsfonts}
\usepackage{multirow}
\usepackage{graphicx,verbatim}
\usepackage{booktabs}
\usepackage{xcolor}
\usepackage{makecell}
\usepackage{float}  
\usepackage{hyperref}
\usepackage{marvosym}
\begin{document}
\title{SynPo: Boosting Training-Free Few-Shot Medical Segmentation via High-Quality Negative Prompts}
\titlerunning{SynPo}
\author{Yufei Liu\inst{1}
\and
Haoke Xiao\inst{2,4}
\and
Jiaxing Chai\inst{1}
\and
Yongcun Zhang\inst{1}
\and
Rong Wang\inst{2}
\and \\
Zijie Meng\inst{3}
\and
Zhiming Luo\inst{1,2}$^{(\href{mailto:zhiming.luo@xmu.edu.cn}{\textrm{\Letter}})}$
}

\authorrunning{Y. Liu et al.}
\institute{Department of Artificial Intelligence, Xiamen University, Xiamen, China \and 
Institute of Artificial Intelligence, Xiamen University, Xiamen, China \and
School of Software and Microelectronics, Peking University, Beijing, China \and
Kuaishou Technology, Beijing, China\\
$^\textrm{\Letter}$Correspondence: \email{zhiming.luo@xmu.edu.cn}\\
Project page: \url{https://liu-yufei.github.io/synpo-project-page}
}

\maketitle              
\begin{abstract}
The advent of Large Vision Models (LVMs) offers new opportunities for few-shot medical image segmentation. 
However, existing training-free methods based on LVMs fail to effectively utilize negative prompts, leading to poor performance on low-contrast medical images. 
To address this issue, we propose \textbf{SynPo}, a training-free few-shot method based on LVMs (e.g., SAM), with the core insight: \textit{\textbf{improving the quality of negative prompts}}.
To select point prompts in a more reliable confidence map, we design a novel Confidence Map Synergy Module by combining the strengths of DINOv2 and SAM. Based on the confidence map, we select the top-k pixels as the positive points set and choose the negative points set using a Gaussian distribution, followed by independent K-means clustering for both sets. Then, these selected points are leveraged as high-quality prompts for SAM to get the segmentation results.
Extensive experiments demonstrate that \textbf{SynPo} achieves performance comparable to state-of-the-art training-based few-shot methods.
\keywords{Few-Shot Segmentation \and Training Free \and Foundation Model.}
\end{abstract}
\section{Introduction}
\begin{figure}
    \centering
    \includegraphics[width=0.98\linewidth]{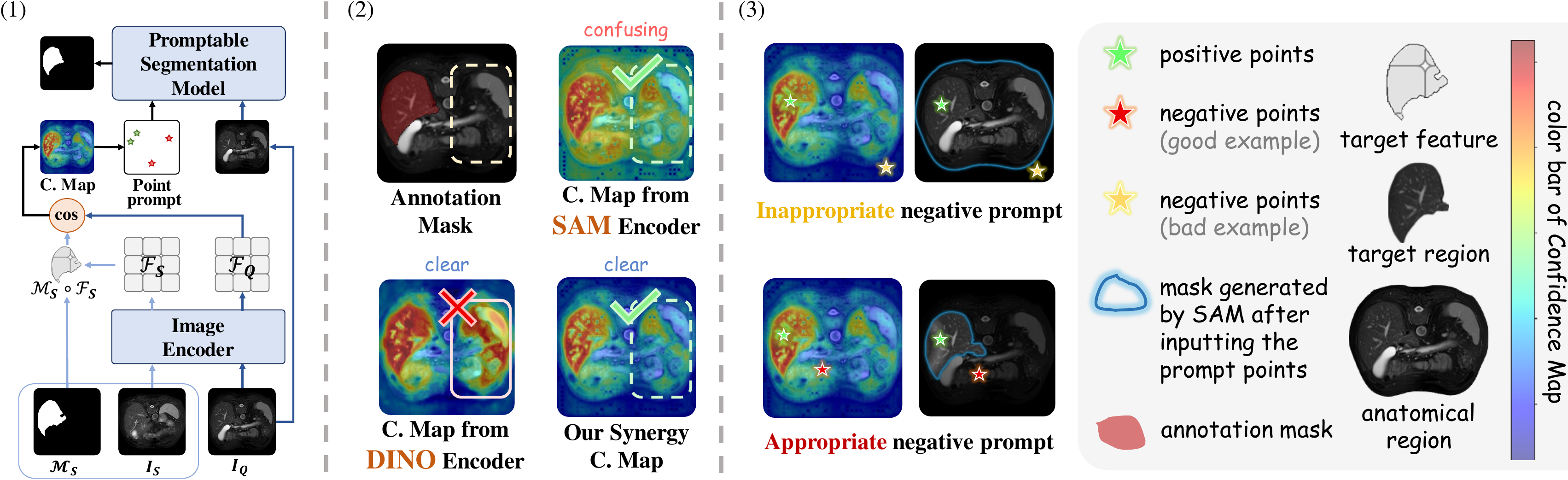}
    \caption{Challenges in Training-Free Few-Shot Segmentation. (1) The universal process of point promptable segmentation model for training-free few-shot. (2) Different Confidence Maps (C. Map) vs. Ground Truth. In the confidence map generated by DINOv2 features, irrelevant regions on the right are mistakenly identified as "similar".
    The confidence map from SAM-ViT features exhibits less clear differentiation in values. 
    Our synergy confidence map leverages the strengths of both while mitigating their respective weaknesses.
    (3) Pilot Experiment. Positioning negative prompts outside the anatomical region, even when using identical positive prompt locations, yields worse segmentation performance compared to placing negative prompts inside. }
    \label{fig:challenge}
\end{figure}

Few-shot medical image segmentation focuses on generating accurate segmentations with a limited number of annotated samples per class or structure.
Recently, few-shot medical image segmentation methods can be classified into two categories: prototype-based methods~\cite{cheng2024few,ouyang2020self,shen2023q,zhu2023few,hansen2022anomaly} and LVMs-based methods~\cite{ayzenberg2024protosam,wang2025osam}, the latter including training-free and fine-tuning methods. 
A representative example of such LVMs is the Segment Anything Model (SAM)~\cite{kirillov2023segment}. 
Whether on natural images or medical images~\cite{tang2025towards,zhang2023personalize,ayzenberg2024protosam}, the process of using SAM for training-free few-shot segmentation task can be summarized in the Fig.~\ref{fig:challenge} (1).
With a support-query image pair and support mask as input, this process firstly extracts feature maps for support-query image pair using a strong pre-trained visual encoder (\textbf{e.g.}, DINO), then computes the 2-D confidence map between query features and the target feature by cosine distance. The maximum value on the confidence map is selected as the point prompt and is used to guide SAM in performing interactive segmentation on the query image.

Although this paradigm achieves good results in the medical field, there is still room for improvement in the generation of high-quality point prompts, which is crucial for it.
First, the utility of confidence maps can still be enhanced.
DINOv2 is widely adopted as the feature extractor in this paradigm~\cite{ayzenberg2024protosam,wang2025osam}, 
excelling at extracting semantic features~\cite{zhang2024tale}. 
However, its employment of relative positional encoding~\cite{vaswani2017attention} compromises the awareness of absolute spatial localization, potentially causing incorrect positive point selection, shown in the bottom left of Fig.~\ref{fig:challenge}(2). 
This impedes the leveraging of anatomical priors for organ positioning, which may result in segmentation's anatomical inconsistency. 
As for SAM-ViT, which is also widely employed, using absolute position encoding with better location awareness can effectively avoid this situation~\cite{dosovitskiy2020image}, as shown in the green box of the image in the top right of Fig.~\ref{fig:challenge}(2). Second, the selection strategy of negative prompts is coarse, resulting in segmentation performance degradation. Current methods identify pixels in the confidence map with minimal similarity to the target region as negative points, causing most negative points to cluster in background areas rather than in the anatomical region. 
The pilot experiment, shown in Fig.~\ref{fig:challenge}(3), demonstrates that this approach is counterintuitive and, therefore, leads to poor results. In medical imaging (\textbf{e.g.}, MRI/CT), such background regions are readily distinguishable from targets, rendering negative prompts redundant for SAM guidance, which leads to over-segmentation, as also noted in~\cite{tang2023can}.

To address these limitations, we propose \textbf{SynPo}, a novel training-free method with  Confidence Map \textbf{Syn}ergy Module and \textbf{Po}int Selection Module. First, drawing from SAM's ability to capture precise low-level spatial information~\cite{kirillov2023segment,ravi2024sam}, which can be a supplement of features from DINOv2, we introduce the Confidence Map Synergy Module. 
This module combines high-level semantic features from DINOv2 with SAM-ViT's absolute spatial cues, leveraging both to enhance anatomical structure capture and refine segmentation boundaries.
Second, the Point Prompt Strategy Module, heuristically selecting negative prompts in the anatomical region, improves the selection of informative negative points, optimizing prompt guidance for segmentation and reducing redundancy prompt information.
Furthermore, we incorporate a Noise-aware Refine Module that utilizes standard morphology and SAM to refine coarse masks. Experiments conducted on both CT and MRI abdominal datasets, focusing on four major organs, demonstrate that our method achieves performance on par with the state-of-the-art few-shot segmentation approaches.

\section{Methodology}
\begin{figure}[t]
   \begin{center}
     \includegraphics[width=\linewidth]{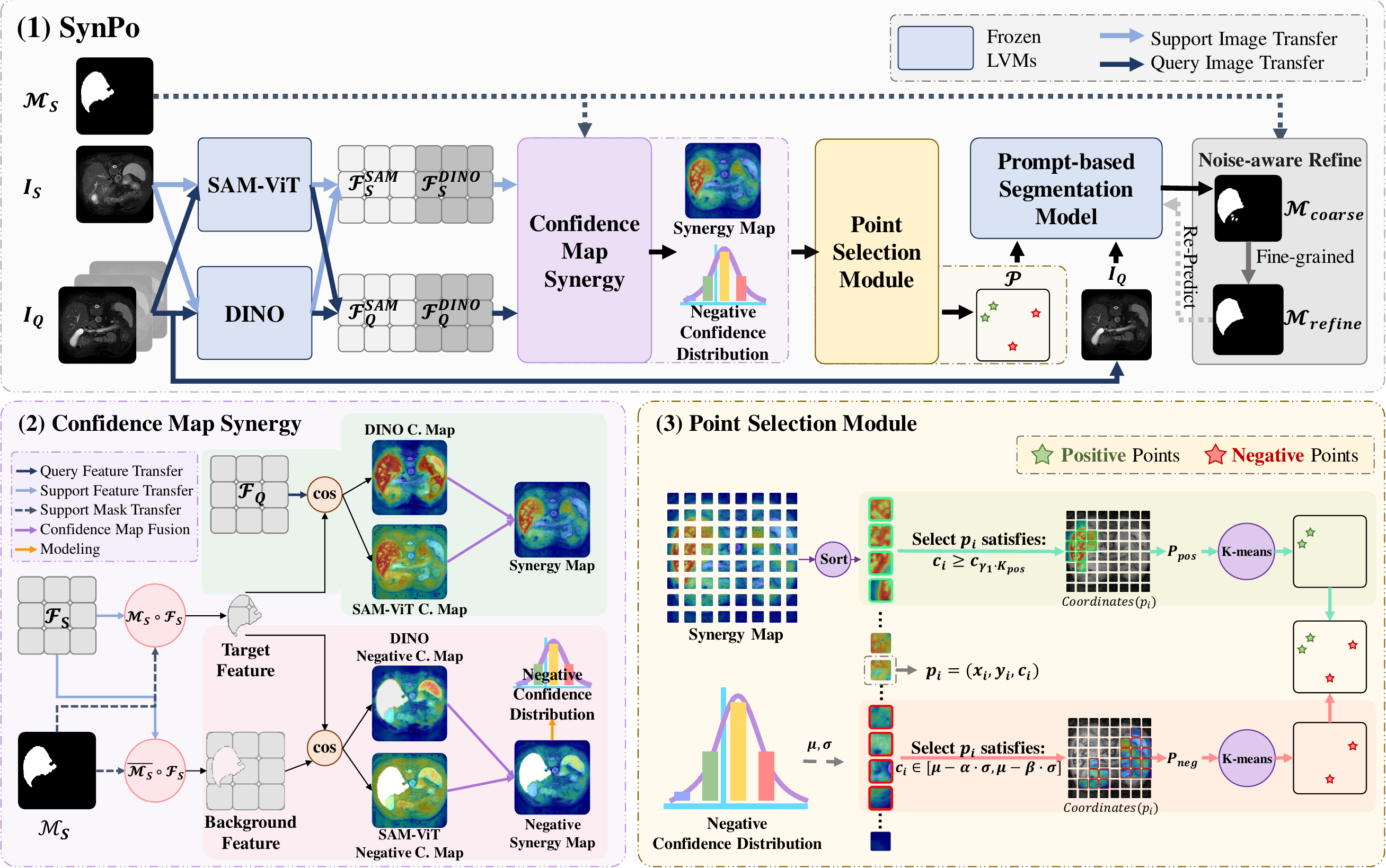}
     \caption{(1)Overview of SynPo Architecture. (2) Illustration of Confidence Map Synergy. (3) Point Selection Module Diagram.}
     \label{fig:pipeline}
   \end{center}
\end{figure}
\subsection{Overview}
SynPo, as shown in Fig.~\ref{fig:pipeline}(1), consists of three key components: Confidence Map Synergy Module (CMSM), Point Selection Module (PSM), and Noise-aware Refine Module (NRM).
Given a support-query pair, first extract zero-shot visual features using pre-trained vision models (SAM-ViT and DINOv2).
In CMSM, the feature maps along with the support mask $\mathcal{M}_S\in \mathbb{R}^{H \times W}$, are used to calculate the synergy map $SynMap\in \mathbb{R}^{H \times W}$ and model the negative confidence distribution $P_{neg}$, which are used to support the generation of prompts.
In PSM, the pixels in the synergy map are sorted by their confidence score into a ranked list, which, along with the confidence distribution, are decisive factors in the selection of point prompts.
Finally, the generated point prompt and Query Image $\textbf{I}_{Q}$ are fed into the SAM to predict the naive mask $\mathcal{M}_{coarse}\in \mathbb{R}^{H\times W}$. Additional NRM is designed for refining the $\mathcal{M}_{coarse}$.

\subsection{Confidence Map Synergy Module}
As shown in Fig.~\ref{fig:pipeline} (2), CMSM is an innovative method with two branches to generate the synergy map, accompanied by the negative confidence distribution as a by-product.
For the upper branch, the support features extracted corresponding to the foreground pixels within the visual concept from $\mathcal{F}_{s}$ using $\mathcal{M}_{s}$:
\begin{equation}
    \left\{\mathcal{T}_S^i\right\}_{i=1}^n = \{x_i|\left. where \right. x_i \in \mathcal{M}_{s} \circ \mathcal{F}_{s} \left. and \right. x_i \ne 0\},
\end{equation}
where $T_S^i\in\mathbb{R}^{1\times c}$ and $\circ$ denotes spatial-wise multiplication. 
Subsequently, we calculate $n$ confidence maps for each foreground pixel $i$ by the cosine similarity between $\mathcal{T}_S^i$ and query feature $\mathcal{F}_Q$:
\begin{equation}
    \left\{\mathcal{S}^i\right\}_{i=1}^n=\left\{\mathcal{F}_{q}\times \left(\mathcal{T}_S^i\right)^\textbf{T}\right\}_{i=1}^n, \quad where\quad \mathcal{S}^i\in\mathbb{R}^{h\times w}.
\end{equation}
Next, we adopt an average pooling to aggregate all $n$ local maps to obtain the overall confidence map $\mathcal{S} \in \mathbb{R}^{h\times w}$ for the target organ. 

For the lower branch generating negative confidence map, we crop the support features of background pixels within the visual concept from $\mathcal{F}_S$ by $\mathcal{\overline{M}}_S$, where $\mathcal{\overline{M}}_S$ represents the inverse of $\mathcal{M}_S$:

\begin{equation}
        \left\{\mathcal{B}_S^i\right\}_{i=1}^{H\times W-n} = \mathcal{\overline{M}}_{s} \circ \mathcal{F}_{s},
\end{equation}
where $\mathcal{B}_S^i \in \mathbb{R}^{1\times c}$ and $\circ$ denotes spatial-wise multiplication. Afterward, we treat $\mathcal{B}_S$ as $\mathcal{F}_q$ and compute the negative confidence map $\mathcal{S}_{neg}$ in the same way. 

Subsequently, we combine $\mathcal{S}^{SAM},\mathcal{S}^{DINO}$ to generate $SynMap$:
\begin{equation}
    SynMap = \delta_{S\text{-}D} \cdot \left(\mathcal{S}^{SAM}\odot \mathcal{S}^{DINO}\right)+\delta_{S} \cdot \mathcal{S}^{SAM}+ \delta_{D} \cdot \mathcal{S}^{DINO},
\end{equation}
where $\odot$ denotes to Hadamard product
and $\delta_{S\text{-}D}+\delta_{S}+\delta_{D}=1$. The first term
captures the non-linear interaction between the two matrices, effectively amplifying extreme values 
in the resultant matrix. This increases the sensitivity to significant deviations in both confidence maps while reducing the impact of more neutral or moderate values. Furthermore, the subsequent terms provide controlled weights to the contribution of each matrix.
Next, the same procedure is applied to $\mathcal{S}_{neg}^{SAM}$ and $\mathcal{S}_{neg}^{DINO}$, resulting in the fused representation, which is then flattened to obtain $SynMap_{neg} \in \mathbb{R}^{(H \times W - n) \times 1}$:
\begin{equation}
    SynMap_{neg} = \text{Flatten}\left(\delta_{S\text{-}D} \cdot \left(\mathcal{S}^{SAM}_{neg}\odot \mathcal{S}^{DINO}_{neg}\right)+\delta_{S} \cdot \mathcal{S}^{SAM}_{neg}+ \delta_{D} \cdot \mathcal{S}^{DINO}_{neg}\right).
\end{equation}
For $SynMap_{neg}$, we model each pixel value $p_i$ in it using a Gaussian probability density function: 
\begin{equation}
    P_{neg}(p_{i}^B|\mu,\sigma^2)=\frac{1}{\sqrt{2\pi\sigma^2}}\exp\left(-\frac{\left(p_i^B-\mu\right)^2}{2\sigma^2}\right),
\end{equation}
where $\mu$ and $\sigma$ are derived through maximum likelihood estimation.
\subsection{Point Selection Module}
We sort each pixel $p_i=(x_i,y_i,c_i)$ in the synergy map by the confidence score of the pixel $c_i$ in descending order, where $x_i,y_i$ are the coordinates of the pixel.

For \textbf{positive} points, we selected the top $\gamma_1 \cdot K_p$ of $p_i$,where $\gamma_1$ is a clustering scaling factor and $K_p$ represents the desired number of positive points. The set of coordinates for the selected points can be written as:
\begin{equation}
    P_{pos} = \left\{Coordinates\left(p_i\right)| i\le \gamma_1 \cdot K_p \right\},
\end{equation}
where $p_i\in \mathbb{R}^2$
Next, we perform K-means clustering on these coordinates and select $K_p$ centroid points, yielding $\mathcal{P}_{pos}$:
\begin{equation}
    \mathcal{P}_{pos} = \text{KMeans}(P_{pos}, K_p),
\end{equation}
where $\mathcal{P}_{pos}\in \mathbb{R}^{K_p\times 2}$ represents the set of coordinates.

For \textbf{negative} points, we select $\gamma_2 \cdot K_n$ pixels at most from the confidence region that are more relevant to the ROI, as defined by the following equation:
\begin{equation}
    P'_{neg}=\{Coordinates(p_i)|c_i \in [\mu - \alpha \cdot \sigma, \mu - \beta \cdot \sigma ]\},
\end{equation}
where the $\mu$ and $\sigma$ derived from $SynMap_{neg}$, $\alpha$ and $\beta$ are user-defined constants that control the bounds of the confidence interval. Subsequently, $\gamma_2 \cdot K_n$ pixels are  randomly selected:
\begin{equation}
R_{neg}=\begin{cases}
    P'_{neg}, ~~&\text{if}~~|P_{neg}|\le \gamma_2 \cdot K_n,\\
    \text{Random}(P'_{neg},\gamma_2 \cdot K_n), ~~&\text{Otherwise}.
\end{cases}
\end{equation}
Then, the clustering similarly is applied to the positive points, yielding $\mathcal{P}_{neg}$:
\begin{equation}
\mathcal{P}_{neg}=\text{KMeans}(R_{neg},K_n).
\end{equation}

Finally, we take the intersection of two sets and assign labels to the points, resulting in the set
\begin{equation}
    \mathcal{P} = \{(x,y,1)|(x,y)\in \mathcal{P}_{pos}\} \cup \{(x,y,0)|(x,y)\in \mathcal{P}_{neg}\},
\end{equation}
which is fed into SAM to generate the segmentation result $\mathcal{M}_{coarse}$.
\subsection{Noise-aware Refine Module}
The module firstly refines the naive coarse mask $\mathcal{M}_{coarse}$ generated by the PSM through an initial erosion operation to remove small noise, followed by a dilation step to restore the main structural regions.
Let $\mathcal{M}_j$ denote the mask obtained for the $\mathcal{M}_{coarse}$'s $j$-th connected region $C_j$, we segment query feature by $\mathcal{M}_j$:
\begin{equation}
    \left\{{T_{Q,C_j}^i}\right\}_{i=1}^{|C_j|} = \mathcal{M}_j \circ \mathcal{F}_Q,
\end{equation}
where $|C_j|$ denotes the total number of pixels in $C_j$. Afterward, we treat
$T_{Q,C_j}$ as $F_q$ and compute the confidence score of $C_j$ in the same way, yeiled $\mathcal{S}_{C_j}$. Thereafter, the mean value is calculated in each connected domain:
\begin{equation}
    score(C_j) = \frac{1}{|C_j|}\mathcal{S}_{C_j}.
\end{equation}

We select the connected region with the highest score as $\mathcal{M}_{refine}$, and feed into the PSM as mask prompts. Combined with the point prompts, this facilitates further refinement of the segmentation. The output is then processed again through the same steps to produce the final segmentation $\mathcal{M}_{final}$.

\section{Experiments and Results}
\subsection{Experimental Settings}
\noindent\textbf{Datasets \& Evaluation Metrics:}
To assess the generalizability of our method, we conducted 1-shot segmentation experiments on Synapse-CT~\cite{landman2015miccai} and CHAOS-MRI ~\cite{kavur2021chaos} datasets, which contains 30 3D abdominal CT scans and 20 3D T2-SPIR MRI scans respectively.
Dataset splits followed the protocols used by ADNet~\cite{hansen2022anomaly} and PerSAM~\cite{zhang2023personalize}. We use the mean Dice score to assess performance, and report the standard deviation of Dice scores across 5-fold cross-validation.

\textbf{Implementation Details:} 
Image preprocessing follows the method described in ~\cite{ouyang2020self}, converting images into
ROI
of size $256\times 256$. 
Features are extracted using DINOv2~\cite{oquab2023dinov2} (model: Sinder~\cite{wang2024sinder}), resulting in a feature map of spatial size $64\times 64$. SAM~\cite{kirillov2023segment} (model: Sam2.1 Hiera large image prediction~\cite{ravi2024sam}) is applied to obtain a feature map of the same spatial size $64\times 64$ for height and width. All experiments are conducted on an NVIDIA RTX-3090. As for Hyper-parameters, $\delta_{S\text{-}D} = 0.8$, $\delta_{S} = 0.1$, and $\delta_{D} = 0.1$ are fixed for Synapse-CT and CHAOS-MRI.

\subsection{Comparison with SOTA Methods}
\begin{table}[t]
\footnotesize
    \caption{Comparison with SOTA Methods. Except PerSAM and SynPo, the data is sourced from their papers.}
    \centering
    \resizebox{\linewidth}{!}{
    \begin{tabular}{clccccccccccc}
    \toprule
     \multirow{3}{*}{Training}&\multicolumn{2}{l}{\multirow{3}{*}{Method}} & \multicolumn{5}{c}{CHAOS-MRI} & \multicolumn{5}{c}{Synapse-CT} \\ 
     \cmidrule(lr){4-8} \cmidrule(lr){9-13} 
        & & & \multicolumn{2}{c}{Upper} & \multicolumn{2}{c}{Lower} & \multirow{2}{*}{Mean} & \multicolumn{2}{c}{Upper} & \multicolumn{2}{c}{Lower} & \multirow{2}{*}{Mean} \\ 
        \cmidrule(lr){4-5} \cmidrule(lr){6-7} \cmidrule(lr){9-10} \cmidrule(lr){11-12}
        & & & Spleen & Liver & LK & RK & & Spleen & Liver & LK & RK &  \\
        \midrule
        \multirow{5}{*}{\textit{w}} & \multicolumn{2}{l}{SSL-ALPNet{$_{\text{\color{gray}{ECCV20}}}$}~\cite{ouyang2020self}}  & 67.02 & 73.05 & 73.63 & 78.39 & 73.02 & 60.25 & 73.65 & 63.34 & 54.82 & 63.02 
        \\
        & ADNet{$_{\text{\color{gray}{MIA22}}}$}~\cite{hansen2022anomaly} &   & 75.92 & 80.81 & 75.28 & 83.28 & 78.82 & 63.48 & 77.24 & 72.13 & 79.06 & 72.97 \\
        & Q-Net{$_{\text{\color{gray}{IntelliSys23}}}$}~\cite{shen2023q}  & & 75.99 &81.74 &78.36 &87.98 &81.02 &74.86 &71.21 &75.26 & 74.79 & 74.03 \\
        & RPT{$_{\text{\color{gray}{MICCAI23}}}$}~\cite{zhu2023few}    & &76.37 &82.86 & 80.72 & 89.82 & 82.44 & 79.13 & 82.57 & 77.05 &72.58 & 77.83 \\
        & GMRD{$_{\text{\color{gray}{TMI24}}}$}~\cite{cheng2024few}    & &76.09  &81.42 &83.96 &90.12 &82.90 & 78.31 & 79.60 & 81.70 & 74.46 & 78.52 \\
        \midrule
        \multirow{5}{*}{\textit{w/o}} & \multirow{2}{*}{PerSAM{$_{\text{\color{gray}{ICLR24}}}$}~\cite{zhang2023personalize}}  & $\mu$ & 69.14 & 42.44 & 64.84 & 71.36 & 61.12 & 65.03 & 65.55 & 58.47 & 60.31 & 62.34 \\
        & &\fontsize{8}{0}\selectfont \textcolor{gray}{$\pm\sigma$} &\fontsize{8}{0}\selectfont \textcolor{gray}{$\pm $8.64}&\fontsize{8}{0}\selectfont\textcolor{gray}{$\pm $3.09}&\fontsize{8}{0}\selectfont\textcolor{gray}{$\pm $2.44}&\fontsize{8}{0}\selectfont\textcolor{gray}{$\pm $5.66} &\fontsize{8}{0}\selectfont\textcolor{gray}{$\pm $3.00} &\fontsize{8}{0}\selectfont\textcolor{gray}{$\pm $9.33} &\fontsize{8}{0}\selectfont \textcolor{gray}{$\pm $3.91} &\fontsize{8}{0}\selectfont\textcolor{gray}{$\pm $4.30} &\fontsize{8}{0}\selectfont\textcolor{gray}{$\pm $4.56} &\fontsize{8}{0}\selectfont\textcolor{gray}{$\pm $1.78}\\
        & \multirow{2}{*}{ProtoSAM{$_{\text{\color{gray}{arxiv24}}}$}~\cite{ayzenberg2024protosam}} & $\mu$ & 76.51 & 81.94 & 71.46 & 81.43 & 77.83 & 65.50 & 87.84 & 69.44 & 71.04 & 73.45 \\
        & &\fontsize{8}{0}\selectfont\textcolor{gray}{$\pm\sigma$} &\fontsize{8}{0}\selectfont\textcolor{gray}{$\pm $7.97} &\fontsize{8}{0}\selectfont\textcolor{gray}{$\pm $6.83} &\fontsize{8}{0}\selectfont\textcolor{gray}{$\pm $6.85} &\fontsize{8}{0}\selectfont\textcolor{gray}{3$\pm $5.16} &\fontsize{8}{0}\selectfont\textcolor{gray}{$\pm $6.85} &\fontsize{8}{0}\selectfont\textcolor{gray}{$\pm $7.72} &\fontsize{8}{0}\selectfont\textcolor{gray}{$\pm $4.71} &\fontsize{8}{0}\selectfont\textcolor{gray}{$\pm $10.21} &\fontsize{8}{0}\selectfont\textcolor{gray}{$\pm $10.52} &\fontsize{8}{0}\selectfont\textcolor{gray}{$\pm $10.21} \\
        &\multirow{2}{*}{SynPo(ours)} & $\mu$ & 80.30  & 77.32 & 77.32 & 83.04 & 81.15 & 83.76 & 81.32 & 75.00 & 79.63 & 79.91 \\ 
        & & \fontsize{8}{0}\selectfont\textcolor{gray}{$\pm\sigma$} &\fontsize{8}{0}\selectfont\textcolor{gray}{$\pm $7.63} &\fontsize{8}{0}\selectfont\textcolor{gray}{$\pm$4.50} &\fontsize{8}{0}\selectfont\textcolor{gray}{$\pm$3.10} &\fontsize{8}{0}\selectfont\textcolor{gray}{$\pm$2.04} &\fontsize{8}{0}\selectfont\textcolor{gray}{$\pm$3.21} &\fontsize{8}{0}\selectfont\textcolor{gray}{$\pm$3.46} &\fontsize{8}{0}\selectfont\textcolor{gray}{$\pm$2.16} &\fontsize{8}{0}\selectfont\textcolor{gray}{$\pm$9.78} &\fontsize{8}{0}\selectfont\textcolor{gray}{$\pm$5.39} &\fontsize{8}{0}\selectfont\textcolor{gray}{$\pm$3.20} \\
        \bottomrule
    \end{tabular}
    }
    \label{tab:sota}
\end{table}

\begin{figure}[t]
    \centering
    \includegraphics[width=\linewidth]{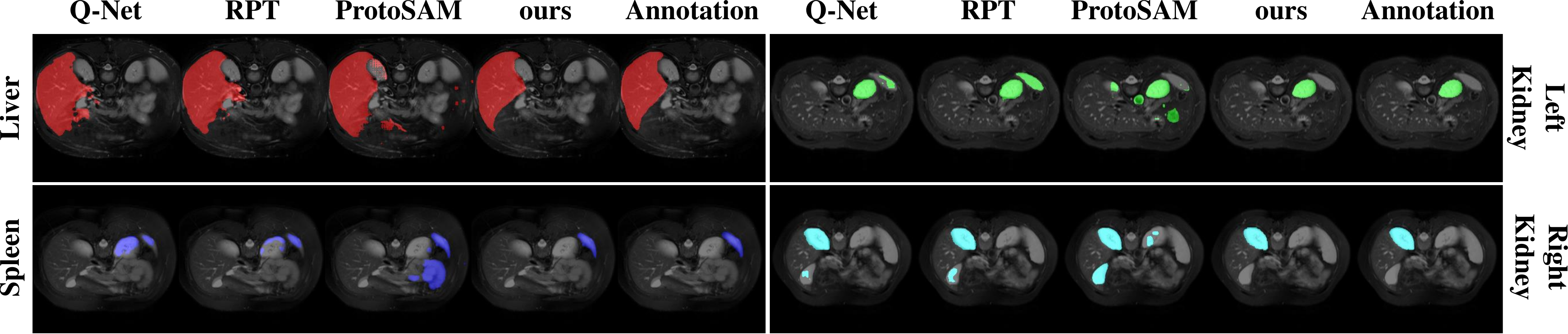}
    \caption{Quality results of different methods.}
    \label{fig:quality_study}
\end{figure}

\textbf{Performance on Synapse-CT.}
As shown in Table~\ref{tab:sota}, SynPo achieves the highest average Dice score (\textbf{79.91}) among training-free methods, surpassing ProtoSAM (73.45) and PerSAM (62.34) by a large margin. Notably, it outperforms some training-based methods in spleen (83.76) and right kidney (79.63), validating the effectiveness of our confidence map synergy and point prompt strategy.

\textbf{Performance on CHAOS-MRI.}
SynPo achieves an average Dice score of \textbf{81.15}, outperforming training-free methods like ProtoSAM (77.83) and PerSAM (61.12), while performing competitively with the top training-based methods like GMRD (82.90). It also excels in spleen segmentation (80.30) and maintains competitive performance on other organs, demonstrating strong generalization.

\textbf{Qualitative Comparison.}
As shown in Fig.~\ref{fig:quality_study}, SynPo achieves segmentation results that closely align with the ground truth. In contrast, Q-Net, RPT, and ProtoSAM tend to produce over-segmented regions. SynPo effectively captures organ boundaries, reduces false positives, and preserves anatomical consistency, benefiting from our Confidence Map Synergy Module and Point Prompt Strategy. These results further demonstrate the superiority of SynPo in few-shot medical image segmentation.

\begin{figure}[t]
    \centering
    \begin{minipage}{0.9\textwidth}
        \centering
        \begin{minipage}{0.5\textwidth}
            \centering
            \begin{table}[H]
                \footnotesize
                \centering
                \caption{ Modules' Ablation Study.}
                \resizebox{\linewidth}{!}{
                    \begin{tabular}{cccc|ll}
                    \toprule
                    \textbf{SAM} & \textbf{DINO} & \textbf{PSM}  & \textbf{NRM} & \textbf{Dice} & \textbf{Gain} \\
                    \midrule
                    \checkmark &   &  &  & 57.48 & - \\
                    \checkmark & \checkmark &  &  & 66.70 & \textcolor{blue}{+9.22} \\
                    \checkmark & \checkmark & \checkmark &  & 78.74 & \textcolor{blue}{\textbf{+12.04}} \\
                    \checkmark & \checkmark & \checkmark & \checkmark & \textbf{79.93} & \textcolor{blue}{+1.19} \\
                    \bottomrule
                    \end{tabular}
                }
                \label{tab:ablation1}
            \end{table}
        \end{minipage}
        \hspace{1em}
        \begin{minipage}{0.45\textwidth}
            \centering
            \begin{table}[H]
                \centering
                \small\caption{Ablation Study of negative prompt strategy on CHAOS.}
                \resizebox{\linewidth}{!}{%
                    \begin{tabular}{l|c} 
                    \toprule
                    Model &  Dice  \\ 
                    \cmidrule(lr){1-2}
                    PerSAM &  $\left. 56.72 \right.$  \\
                    +Negative Point Selection &  $\left. \textbf{68.90} \right. $ \\
                    \bottomrule
                    \end{tabular}
                }
                \label{tab:ablation2}
            \end{table}
        \end{minipage}
    \end{minipage}

    \begin{minipage}{\textwidth}
        \centering
        \includegraphics[width=0.8\linewidth]{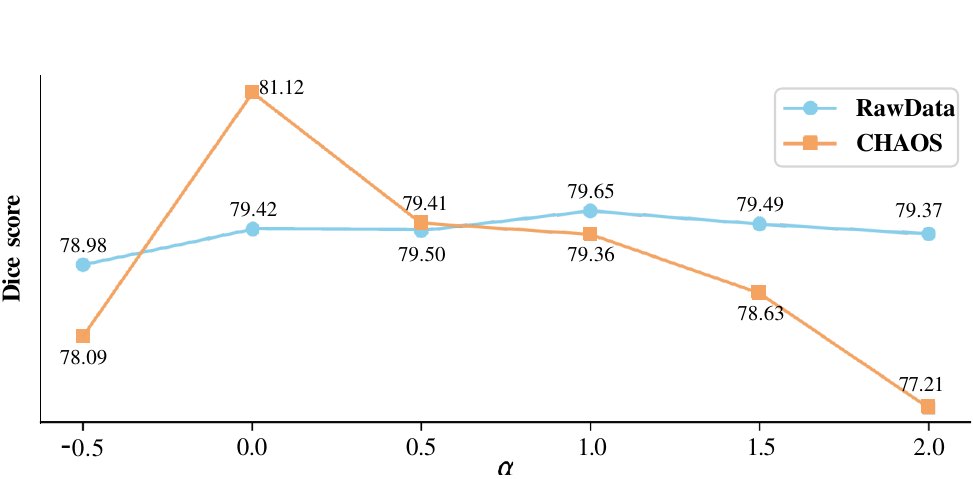}
        \caption{Parameter Experiment of $\alpha$ \& $\beta$, $\beta=\alpha-1.5$.}
        \label{fig:parameter}
    \end{minipage}
\end{figure}

\subsection{Ablation study}

To evaluate the contribution of each module, we conducted an ablation study on the Synapse-CT dataset (Table~\ref{tab:ablation1}). The results show that adding DINO to SAM-ViT improves performance by \textbf{+9.22}, and further incorporating the PSM provides an additional \textbf{+12.04} gain. The NRM further boosts performance by \textbf{+1.19}, demonstrating the effectiveness of our synergy-based design.

Additionally, we replaced the negative point selection strategy in PerSAM to evaluate its impact on CHAOS-MRI (Table~\ref{tab:ablation2}). This replacement yields a significant Dice improvement of \textbf{+13.18}, highlighting the importance of effective negative prompt selection in transferring natural image knowledge to medical.

\textbf{Parameter Experiments} regarding the confidence interval bounds controller $\alpha$ and $\beta$ are also conducted, shown in Fig.~\ref{fig:parameter}. The x-axis represents the value of $\alpha$ (with $\beta=\alpha-1.5$) and the y-axis denotes the Dice score. The figure shows that
in CHAOS and RawData, the highest dice score was achieved under the confidence interval $[\mu,\mu+1.5\sigma]$ and the confidence interval $[\mu-\sigma,\mu+0.5\sigma]$.
The figure proves low Dice scores at high-value $large$, validating our proposed negative point selection strategy that prioritizes not least but less similar points.

\section{Conclusion}
Our model SynPo addresses the limitations such as anatomical consistency of existing training-free few-shot medical segmentation methods by improving negative prompt quality through a Confidence Map Synergy Module and a Point Selection Module. Experiments on CT and MRI datasets demonstrate that SynPo achieves performance comparable to state-of-the-art methods and effectively transfers knowledge from LVMs trained on natural images to the medical image domain. Our work highlights the importance of high-quality negative prompts for few-shot medical image segmentation.

\begin{credits}
\subsubsection{\ackname} This work is supported by the National Natural Science Foundation of China (No.~62276221), the Fujian Provincial Natural Science Foundation of China (No.~2022J01002).
Yufei Liu would like to express her deep gratitude to her late maternal grandfather Chuanwei Lei, whose memory remains a source of strength.
\end{credits}
\bibliographystyle{splncs04}
\bibliography{MICCAI2025}
\end{document}